# A Chinese Multi-label Affective Computing Dataset Based on Social Media Network Users

Jingyi Zhou[1,2], Senlin Luo[1], Haofan Chen[3]

1 School of Information and Electronics, Beijing Institute of Technology, Beijing, 100081, China
2 Institute of Scientific and Technical Research on Archives, Beijing, 100050, China
3 China Electronics Engineering Design Institute Co., Ltd., Beijing, 100142, China
corresponding authors: JingyiZhou (annezjy94@163.com), Senlin Luo (luosenlin2019@126.com)

## Abstract

Emotion and personality are central elements in understanding human psychological states. Emotions reflect an individual's subjective experiences, while personality reveals relatively stable behavioral and cognitive patterns. Existing affective computing datasets often annotate emotion and personality traits separately, lacking fine-grained labeling of micro-emotions and emotion intensity in both single-label and multi-label classifications. Chinese emotion datasets are extremely scarce, and datasets capturing Chinese users' personality traits are even more limited. To address these gaps, this study collected data from the major social media platform Weibo, screening 11,338 valid users from over 50,000 individuals with diverse MBTI personality labels and acquiring 566,900 posts along with the users' MBTI personality tags. Using the EQN method, we compiled a multi-label Chinese affective computing dataset that integrates the same users' personality traits with six emotions and micro-emotions, each annotated with intensity levels. Validation results across multiple NLP classification models demonstrate the dataset's strong utility. This dataset is designed to advance machine recognition of complex human emotions and provide data support for research in psychology, education, marketing, finance, and politics.

## Background & Summary

Text serves as a crucial medium for human emotional expression, and the primary task of text-based affective computing is to extract, analyze, understand, and generate subjective information from textual content. Over the past three years, text-based affective computing has seen rapid advancementsand applications across numerous fields, including psychology[1,2,3], education[4,5], public welfare[6], marketing[7], finance[8,9], politics[10], creative text generation[11], robustness of large models[12], and bias analysis[13]. Emotion detection has also become a focal point in chatbot research, as seen with the release of Microsoft's empathetic social chatbot in 2020[14]and Li's empathy-enabled chatbot with emotion causality[15].

Most emotion research relies on Ekman and Plutchik's emotion models[16], where binary classification denotes the presence or absence of specific emotions, and ternary classification often uses categories such as positive, neutral, and negative. Although these models are straightforward and targeted, they struggle to capture the complexity of human emotions, frequently failing to identify the nuanced and varied emotions embedded in text (termed micro-emotions). A shift towards multi-category, fine-grained, and continuous annotation models has thus become an inevitable trend in text-based affective computing. The MuSe 2022 Challenge[17] introduced humor and emotional stress as dimensions, extending emotion models with additional states for customized emotional representation. ChineseEmoBank[18] provides fine-grained analysis of emotion intensity and is the first Chinese resource offering multi-level granularity in affective dimensions. In 2020, Google's team launched GoEmotions[19], a dataset containing 28 emotion categories with 58,009 manually annotated text samples: single-label annotations account for 83%, double-label for 15%, triple-label for 2%, and 4-5 labels for 0.2%. GoEmotions pioneered a highly detailed multi-category, multi-label micro-emotion dataset; however, it offers limited multi-label annotations and lacks intensity differentiation among tags or personality traits.

Emotion datasets primarily depend on manual annotation, which is costly and prone to subjective bias, particularly in the lack of micro-emotion labeling with intensity levels, limiting machine comprehension of subtle human emotions. Our previous research[20] introduced a micro-emotion annotation framework with continuous intensity scores, and this paper applies that framework to annotate micro-emotion data, establishing the Chinese Multi-label Affective Computing Dataset (CMACD) based on social media network users.

Emotion detection primarily focuses on the emotions conveyed in text, often overlooking the stable internal factors of the text creator, namely personality. Personality, an essential concept in psychology, encompasses enduring response patterns toward others, events, and oneself, shaping an individual's unique adaptive abilities in social contexts. It reflects aspects of integrity, stability, uniqueness, and sociality within one's psychological makeup[21]. Numerous psychological studies emphasize personality as a critical psychological indicator and extend its application from personality research to affective computing. For instance, Kashani et al.[22] assessed the impact of personality on emotional states in video game communication, Kamran et al.[23] applied personality traits to explain emotions in short texts, and Kitkowska[24] investigated how emotions and personal traits influence interactions with privacy policies. Personality-based affective computing has also garnered attention within large model research[25]. Since April 2022, the search index for the Myers-Briggs Type Indicator® (MBTI) on Baidu has surged, and in 2023, it topped the most popular terms on Weibo, reflecting sustained public interest in MBTI-related topics. Widely regarded as one of the most popular and reliable personality prediction methods[2], MBTI has inspired significant attention. However, a publicly available dataset that integrates personality traits and emotions remains absent, and our CMACD dataset fills this gap. This dataset annotates each user's Weibo post with both personality traits (MBTI) and emotions, along with emotion intensity values.

CMACD is a Chinese dataset focusing on personality and emotion. A user with a unique personality will express rich macro and micro emotions in their posts. As shown in Fig. 1.

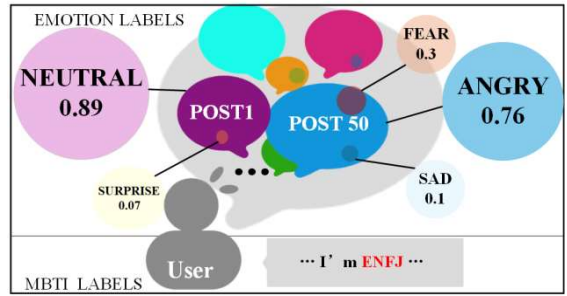

Fig. 1.Miniature of a user in CMACD. it is a snapshot of a user in CMACD, where the numbers in the Fig. represent the intensity of emotions.

Chinese, one of the six official languages of the United Nations, is spoken by approximately 1.39 billion people worldwide, comprising nearly 20% of the global population[26]. China has 1.092 billion internet users, with over 600 million active users on Weibo[27,28]. Numerous studies on text-based affective computing in Chinese social networks have emerged. For example, Wang Haoet al.[29] and Wang Youwei et al.[30] conducted research on misinformation detection, Kong et al.[31] and Tan et al.[32] explored psychological and medical issues, and Li Tiejun et al.[33] and Xiong et al.[34] developed recommendation systems based on sentiment analysis. Other studies have focused on topics such as the "Double Reduction" policy[35], the "Wandering Elephants" event[36], flood disasters[37], traditional Chinese medicine[38], and COVID-19[39, 40]. Most emotion datasets used in these studies are small, privately collected Weibo datasets that are not publicly available. In a few cases, researchers have used English datasets limited to positive, negative, and neutral classifications. Publicly available Chinese emotion datasets are extremely scarce; the authors identified only one high-quality six-class single-label Chinese emotion dataset, primarily used for emotion detection competitions[41]. Given the nuanced nature of Chinese, along with its linguistic diversity, flexible grammar, and deep cultural roots, affective computing research involving Chinese cannot fully rely on English datasets.

In summary, current Chinese emotion datasets face multiple limitations: the scarcity of publicly available datasets, the lack of quantified emotion intensity data, the predominance of single-label annotations, the absence of datasets with Chinese personality traits, and the rarity of publicly available Chinese multi-label emotion datasets.CMACD addresses these shortcomings by offering the first Chinese dataset that integrates personality traits with multi-level quantifications of emotion intensity. This dataset is of significant reference value for advancing machine understanding of human emotions and supporting research in psychology, education, public welfare, marketing, finance, and politics.

The primary contributions of this paper are as follows:

1. CMACD is the first dataset to unify personality and emotion within a single dataset.

2. CMACD represents the first large-scale dataset of Chinese personality traits based on online user data.

3. CMACD, utilizing our published automatic emotion detection annotation tool, is the first to achieve machine-labeled macro-emotions and micro-emotions.

# Methods

The CMACD dataset encompasses four key dimensions: users, user Weibo texts, MBTI personality types, and six emotions with intensity scores, forming a large-scale Chinese dataset. The components of users, user Weibo texts, and MBTI personality types are drawn from publicly available personality information on Weibo, where users either discuss MBTI or self-identify with certain personality traits. This dataset has been carefully compiled in strict compliance with user privacy protections. The six emotion labels with intensity levels are derived from an existing publicly annotated single-label Weibo emotion dataset[41] and are labeled using the Extended Quantification Network (EQN) framework based on a BERT model.The workflow of our study is illustrated in Fig. 2.

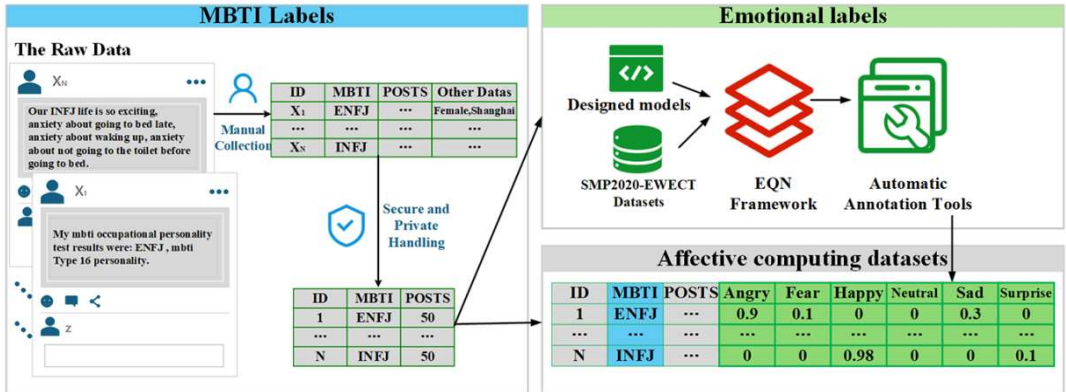

Fig. 2.Workflow for Creating the CMACD Dataset. As shown in Fig. 2, the main process of creating the CMACD dataset involves: manually collecting users with self-identified MBTI personality types and their posts, followed by data filtering and cleaning (including user privacy protection) to create the MBTI dataset. Then, a model, pre-trained using the EQN framework on a high-quality manually annotated emotional Weibo dataset, is applied to label each Weibo post in the MBTI dataset with multiple emotion intensities.

---

1. **Identification of MBTI Users**

Traditional manually labeled datasets require significant human resources, and annotating MBTI personality traits necessitates substantial involvement from psychology professionals. To ensure reliability in MBTI personality trait data, we selected users who had verified their MBTI traits through self-confirmation. To maintain dataset validity, we also established minimum posting requirements for users included in the dataset.

(1) Large-Scale Search (Initial Screening). Through extensive manual searches, we identified Weibo users who publicly self-identify with an MBTI type. Using fuzzy search methods, we collected data from 51,000 users by searching phrases like "I am an INTJ" for each of the 16 personality types, amassing over 5 million posts.

(2) Manual Review (Detailed Screening). To accurately identify users with a specific personality type, we conducted a manual review. For instance, when reviewing users with an ISTJ personality, we included affirmative phrases like "I am an ISTJ," "ISTJ certified," "ISTJ type," "as an ISTJ," "we ISTJs," "true ISTJ," and "100% ISTJ." Users employing such expressions in their posts were selected for further personality verification to ensure the accuracy of MBTI labels.

(3) Expanding Manual Filter Range (Further Detailed Screening). To capture more users with specific personality traits, we compiled a list of popular colloquial expressions associated with each MBTI personality type, as shown in Table 1. Most of these terms are anthropomorphic labels provided by official MBTI testing sites and served as critical reference points in our manual screening process.

Table 1. Popular Online Expressions Corresponding to Different MBTI Personality Types.

| INTJ | INTP | ENTJ | ENTP |
|---|---|---|---|
| Purple Old Man, Coffee Cup | Potion Sister, Airpods Sister, Small Fries | Purple Big Sister, Purple Eldest Sister, Overbearing Boss | Broken Eyebrows, Box Spirit, Amazon Box |
| **INFJ** | **INFP** | **ENFJ** | **ENFP** |
| Magic Old Man, Green Old Man | Little Butterfly, Crying Cat Head | Great Sword | Agent, Happy Puppy |
| **ISTJ** | **ISFJ** | **ESTJ** | **ESFJ** |
| Blue Old Man, Sandwich | Little Nurse, Mother | Blue Big Sister, Ruler Sister | Little Cake, Umbrella Brother |
| **ISTP** | **ISFP** | **ESTP** | **ESFP** |
| Electric Drill Man, Repairman | Little Painter, Durian Head | Sunglasses, Blind Person | Sand Hammer Sister, Little Sand Hammer |

Exclusion of Bot Users. Weibo contains a significant number of bot-generated posts, which lack relevance to affective computing analysis. These bot users were removed from the aggregated user list.

(5) Manual Collection of Post Data by User ID. We created 16 personality folders named according to each MBTI type. Posts collected from each user were stored in CSV files named by their user ID, with each row representing a Weibo post. These CSV files were placed in the corresponding personality folder based on the user's self-identified MBTI type.

**2. Data Preprocessing**

User activity levels on Weibo vary significantly, with some users posting thousands of tweets within a given timeframe, while others post only a few. Individual posts may range from thousands of characters to just a single word or emoji. Given that tweets often contain substantial emotional content, we implemented the following preprocessing steps to ensure dataset quality:

(1) Removed all URLs from the dataset, as they offer no clear information about emotional polarity.

(2) Deleted posts containing promotional or advertisement content.

(3) Replaced any personal names in posts with the placeholder "[name]" and removed various numeric identifiers to protect user privacy.

(4) Deleted posts shorter than 30 characters or longer than 150 characters.

(5) Excluded users with fewer than 50 posts following the above preprocessing steps.

Ultimately, we selected 11,338 users, randomly sampling 50 posts from each. Each post ranged from 30 to 150 characters in length, resulting in a total of 566,900 posts, each categorized under one of the 16 MBTI personality types.

**3. Labeling Emotions with Intensity Scores Using the EQN Framework**

We applied emotion labeling to the organized MBTI dataset. Currently, multi-label classification datasets predominantly rely on manual annotations—a time-consuming and costly process. Furthermore, manual annotation is prone to subjectivity, which can lead to errors or omissions in the labels. Human emotional expression varies in intensity, yet existing emotion labeling methods only categorize emotions without quantifying their strength, resulting in a lack of detail. However, machine learning techniques can capture more subtle features, enhancing the model's ability to classify emotions and deepen the understanding of human emotions.

To address these issues, we implemented machine-driven multi-label emotion annotation with intensity scoring at the sentence or paragraph level. The implementation process is illustrated in Fig. 3.

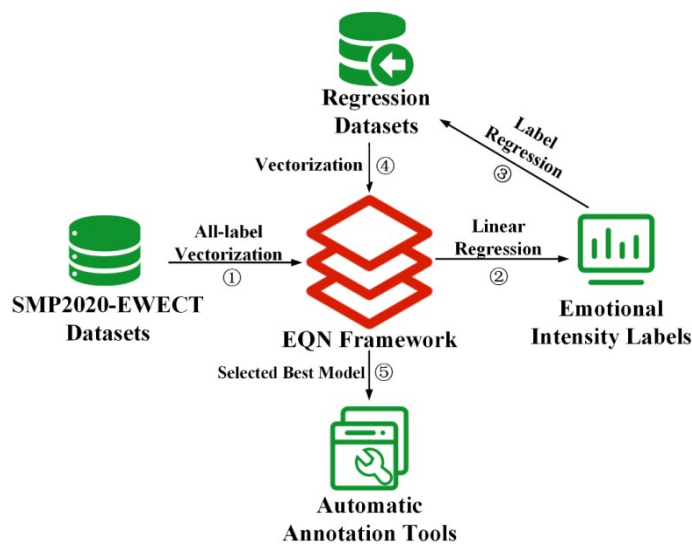

Fig. 3 Implementation Process of Multi-Label Emotion Annotation in CMACD. As shown in Fig. 3, we began by selecting the high-quality SMP2020-EWECT Dataset, a single-label dataset with six emotions (angry, fear, happy, neutral, sad, surprise). Label values were initialized using a full-label mapping approach. After processing this data through the model in the EQN framework, emotion labels were applied to the original training set. A regression adjustment was then performed on the annotated dataset, setting values of originally labeled emotions to 1 and retaining EQN-generated values for previously unlabeled data. The model within the EQN framework was retrained on this regressed dataset, which significantly improved its detection performance (in our BERT-based EQN model, accuracy rose from 78.3% to 81.2% after the second training). Finally, this enhanced model was used to label each post in the MBTI dataset with multi-label emotions and intensity scores, forming the CMACD dataset.

- - - - - - - - - - - - - - - - - - - - - - - - - - - - - - - - - - - - - - - - - - - - - - - - - - - - -

The structure diagram of dataset annotation using the retrained model we designed is shown in Fig. 4.

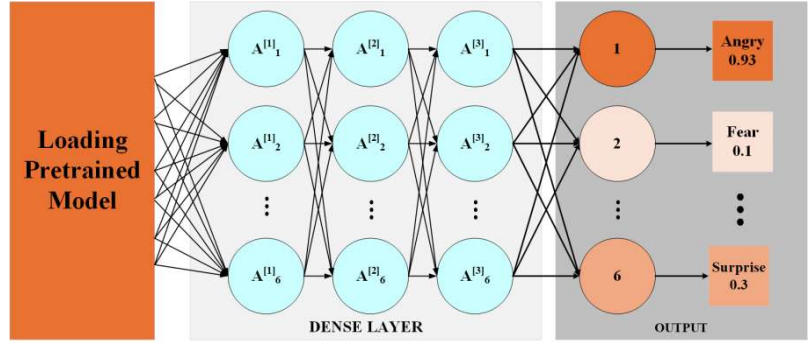

Fig. 4. Emotion Annotation Structure in CMACD. First, the trained model is loaded, and its output is connected to a fully connected neural network (also known as a dense layer), which finally produces an output with intensity values corresponding to the six emotion categories.

For further details, refer to our paper, "Extended Quantification Network: An Efficient Micro-Emotion Detection and Annotation Framework"[20].

The dataset is described in three parts: analysis of the MBTI data, the six emotion categories, and a statistical analysis of the integrated MBTI personality and emotion data.

## 4. MBTI Data Analysis

### (1) Distribution of Users across the 16 Personality Types

As shown in Fig.5, the distribution of users across the 16 MBTI personality types in the CMACD dataset is depicted. The number of users for each personality type is indicated above the corresponding bars in the histogram. It can be observed that the number of users varies significantly across different personality types. Users of the INFP, ENFP, and INFJ types each exceed 1,000, while users of the ESTP, ESTJ, ESFJ, and ENTJ types number fewer than 300. Among these, the number of INFP users is 12.46 times greater than that of ESTP users.

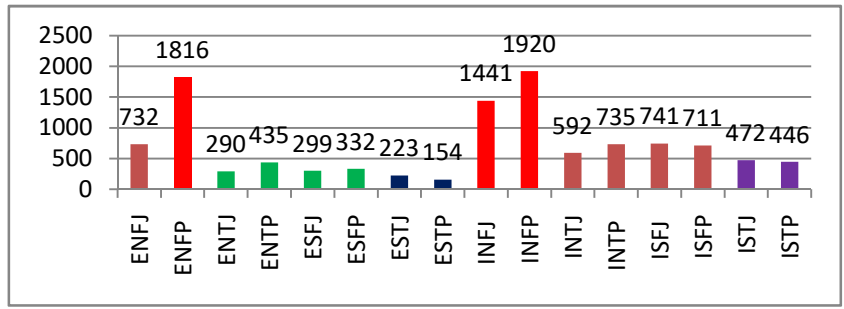

Fig. 5 Distribution of Users Across 16 Personality Types.

### (2) Word Count Statistics for Each Post Across the 16 Personality Types

In the entire dataset, 50 posts were recorded for each user, with an average post length of 63.7 characters. The longest post contains 145 characters, while the shortest contains 35 characters. Fig. 6 displays the average word count per post across the 16 personality types. The calculation method involves dividing the total word count of all posts in a specific personality type by the number of users of that type, and then dividing by 50. The chart indicates that there are significant differences in the average post length among the different personality types, despite minimal variation in the minimum word count and the overall average.

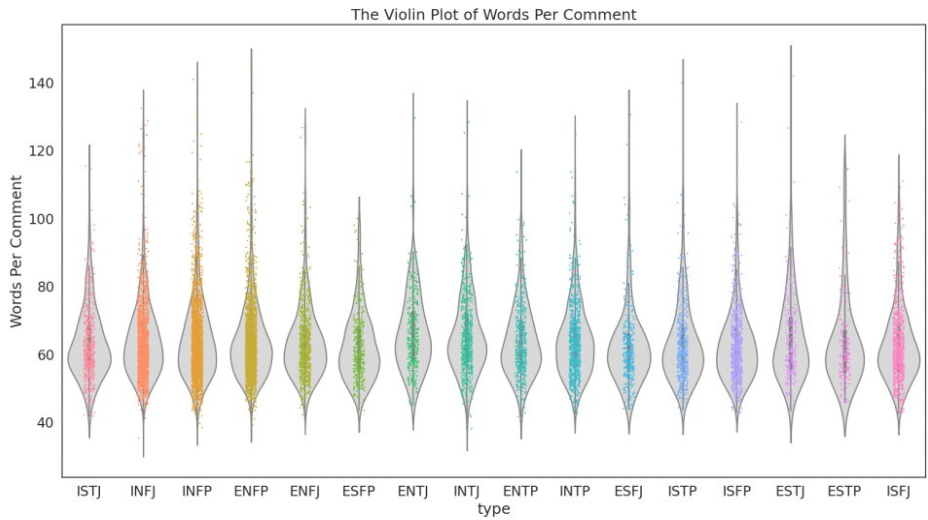

Fig. 6 Distribution of word count per post for the 16 personality types. Fig. 6 illustrates the word count distribution for each post across different personality types. From the Fig., it is evident that there are distinct group differences in the word count distribution of posts among the various personality types.

---

**(3) Distribution of Users Across the Four Axes and Eight Polarities of MBTI**

According to the MBTI personality classification theory, personality traits are analyzed from four perspectives: cognitive functions, information processing, decision-making styles, and lifestyle orientations. These correspond to the four dimensions of IE, NS, TF, and JP (referred to as the "four axes" in this study). Each dimension is further divided into two opposing tendencies (for example, IE is divided into I and E), resulting in the "four axes and eight polarities" that combine to form the 16 personality types. Table 2 and Fig. 7 present the distribution of users across these four axes and eight polarities in the dataset.

| Four-Axis | Polarity | Number of Users | Polarity | Number of Users |
|---|---|---|---|---|
| IE | Introversion (I) | 7058 | Extroversion(E) | 4281 |
| NS | Intuition (N) | 7961 | Sensing(S) | 3378 |
| TF | Thinking (T) | 3347 | Feeling(F) | 7992 |
| JP | Judging (J) | 4790 | Perceiving(P) | 6549 |

**Table 2.Distribution of users of MBTI four-axis eight-polarity.**

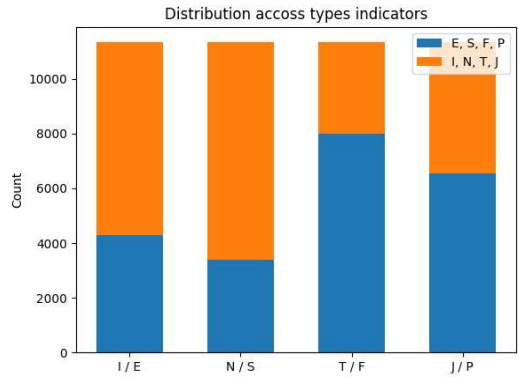

Fig. 7. Four-axis octupole user distribution.

## 5. Quantitative Emotional Data Analysis

Using the previously developed EQN framework[20], we fine-tuned a BERT pre-trained model on a manually annotated dataset consisting of six emotion categories to train the annotation model[42]. This trained model provides multi-label emotional annotations and intensity scores for each post in the dataset. (The trained annotation model and labeling code can be accessed at: https://github.com/yeaso/Expansion-Quantization-Network-EQN). The final dataset includes 566,950 posts from 11,338 Weibo users, covering six emotions (anger, fear, happiness, neutrality, sadness, and surprise) with quantitative multi-label annotations. Fig. 8 illustrates an example of this quantified emotional data.

Fig. 8. Quantitative annotation of sentiment labels. The emotional annotations in the dataset are assigned intensity levels ranging from 0 to 1 (where 0 indicates absence and 1 denotes maximum intensity). Each post may contain multiple emotions. Our EQN annotation tool utilizes a full-label quantization method to effectively capture the correlations and dependencies between different labels. This allows us to annotate intensity scores that reflect both macro and micro emotional aspects in the text. By setting a threshold, we define valid labels—scores above this threshold are considered valid annotations. In our experiment, this threshold is set at 0.05; any emotional intensity below 0.05 is uniformly set to 0. Labels with an intensity of 0 indicate that the post does not exhibit that particular emotion.

For a more detailed analysis of the quantified emotional data, we provide a comparative analysis with the MBTI data, identifying similarities and differences to better understand and analyze the relationship between MBTI and emotional annotations. Please refer to the section "MBTI and Emotion Data Analysis" below.

## 6. MBTI and Emotion Data Analysis

For clarity, the following variables and formulas are defined:

F(i): The intensity score of each post for each emotion category, with values ranging from (0,

1). i= [angry, fear, happy, neutral, sad, surprise].

$$F(i) = \begin{cases} F(i) & \text{if } F(i) \geq t \\ 0 & \text{if } F(i) < t \end{cases} \quad (1)$$

Here t is the threshold defined for the annotation label.

Pnum: Number of posts per personality type.

$$Pnum = \sum F(i) \quad (2)$$

Enum（i）: The number of energy level scores of each type of emotion for each type of personality>= t. (This data set sets t=0.05).

Sume（i）: Total energy. The sum of the energy level scores of each emotion type for each personality type.

$$Sume(i) = \sum_{1}^{Enum} F_i \quad (3)$$

LAE(i): Local average energy. The average amount of energy a certain type of emotion possesses within each personality type.

$$LAE(i) = \frac{Sume(i)}{Enum(i)} \quad (4)$$

GAE(i): Global average energy. The average amount of emotion contained in posts by each personality type.

$$GAE(i) = \frac{Sume(i)}{Pnum} \quad (5)$$

ANL（i）: The average number of labels. The probability of occurrence of each type of emotion for each type of personality.

$$ANL(i) = \frac{Enum(i)}{Pnum} \quad (6)$$

QEL(i): The quantity exceeding the local average energy. The number of posts for each personality type with emotion levels exceeding LAE(i) for each emotion category.

$$QEL(i) = \frac{Enum(i)}{Pnum} - \frac{Sume(i)}{Enum(i)} \quad (7)$$

QEG(i): The quantity exceeding the global average energy. The number of posts for each personality type with each emotion category exceeding GAE(i) levels.

$$QEG(i) = \frac{Enum(i) - Sume(i)}{Pnum} \quad (8)$$

LHEP(i): Local high expression probability. The local high-expression probability for each emotion within each personality type.

$$LHEP(i) = \frac{QEL(i)}{Pnum} \quad (9)$$

GHEP(i): Global high expression probability. The local high-expression probability for each emotion across posts within each personality type.

$$GHEP(i) = \frac{QEG(i)}{Pnum} \quad (10)$$

The formulas (1)–(5) compute the total and average emotional intensity for each emotion across the 16 personality types, reflecting the strength of each type's emotional expression.

Formulas (6)–(10) calculate the probability of users across the 16 personality types expressing the six emotions. Using these definitions, we performed corresponding statistical analyses on our emotion dataset, with the results presented in Supplementary Table 1.

To facilitate observation and analysis, we have created several charts based on the data in Supplementary Table 1 to visually represent the various data categories.

A key feature of our multi-label emotional dataset is that each annotated label is associated with a specific intensity value. With these quantified data, we are able to analyze the distribution of emotional intensity across different personality types, providing valuable insights into the correlation between personality and emotion.

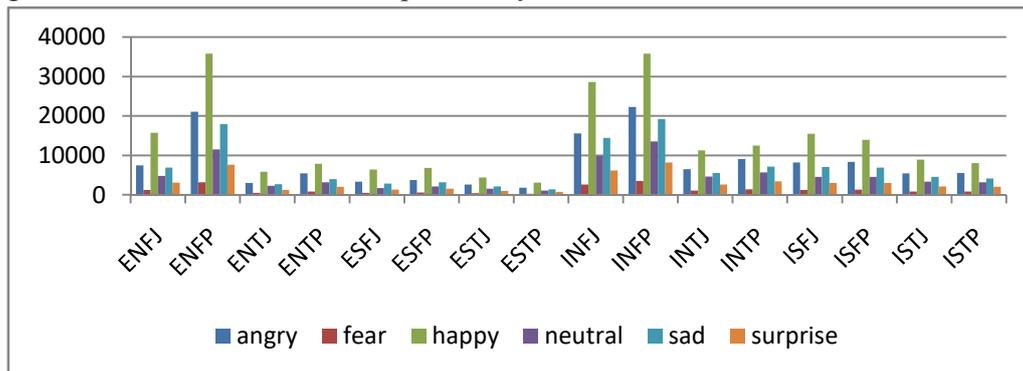

Fig. 9. Overall distribution of the number of people with six emotions in the 16 personality types. Fig. 9 presents the overall distribution of six emotions across the 16 personality types, based on the Sume data from Supplementary Table 1.

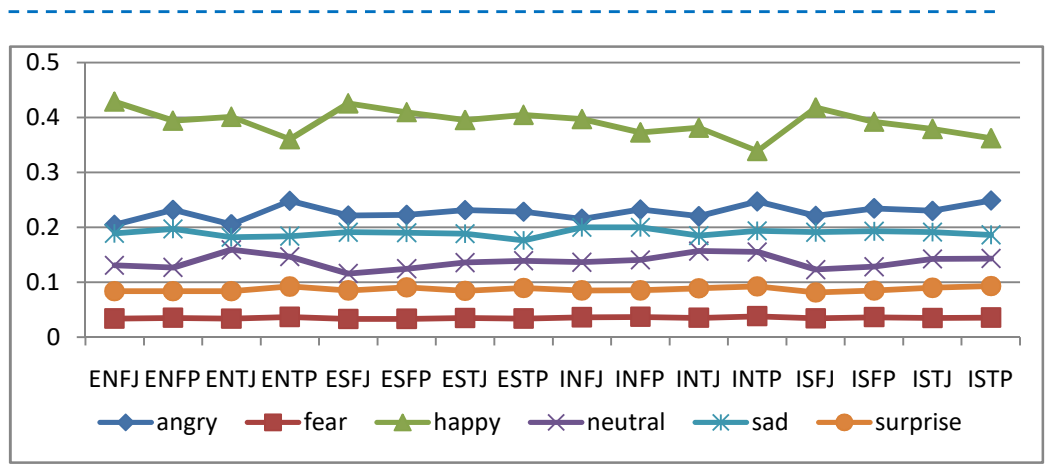

Fig. 10. Distribution of average energy values for all users across the 16 personality types.

Based on the GAE data in Supplementary Table 1, the average energy values for emotions across the 16 personality types display the mean energy for each emotion. As shown in the chart, the average energy values for the emotions, in descending order, are: happiness, anger, sadness, neutrality, surprise, and fear. Notably, the energy values for surprise and fear are relatively low, indicating an overall stability in emotional expression among these personality types. Happiness has the highest energy level, suggesting that individuals across all personality types express happiness with the greatest intensity. Specifically, ENFJ, INFJ, ESFJ, and ISFJ types exhibit the strongest expressions of happiness and the least expressions of anger, highlighting an optimistic emotional tendency within these groups. In contrast,

ENFP, ENTP, and INTP types show a more pronounced expression of anger, while their happiness expression is relatively restrained compared to other personality types.

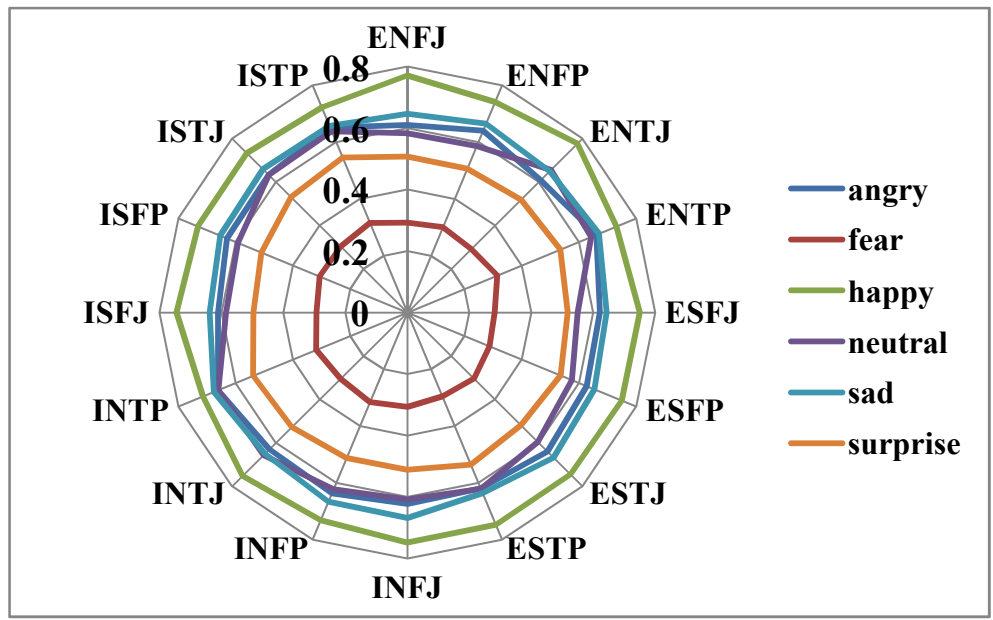

Fig. 11. Probability Distribution of Each Emotion Across Personality Types. Based on the ANL data from Supplementary Table 1, this Fig.11 illustrates the probability distribution for each emotion across the different personality types. Although the occurrence probabilities for various emotions show slight differences among the 16 personality types, "happy" has the highest occurrence probability across all user groups, while "fear" and "surprise" consistently have the lowest and second-lowest probabilities, respectively.

**Data Records**

The CMACD dataset comprises data from 11,338 Weibo users, with each user contributing 50 posts. Each post is limited to 30–150 characters, totaling 566,900 posts. The dataset includes users' MBTI personality types along with a comprehensive affective computing dataset containing six emotion intensity values. The data records consist of the source of the dataset, the dataset used for training the annotation model, and the format for data storage.

**1. Source of Collected Dataset**

The user Weibo corpus and MBTI personality characteristics used for this dataset were collected from the Weibo platform, accessible at https://weibo.com/.

**2. Training Dataset for Emotion Annotation**

The emotion intensity annotations for this dataset were performed using the EQN emotion annotation framework. The training dataset used for the framework is the "SMP2020 Weibo Emotion Classification Technical Evaluation General Weibo Dataset." This dataset contains 27,768 Weibo posts, each annotated manually with one of six emotional categories, making it a single-label dataset. The dataset can be downloaded at https://smp2020ewect.github.io.

**3. Storage Structure of the CMACD Dataset**

The CMACD dataset is organized in a hierarchical structure, with folders named according to the 16 MBTI personality types, as shown in Fig. 12. Each folder contains CSV files corresponding to individual users classified under that personality type. In each user's CSV file, 50 rows represent the user's posts, with columns containing the post content and scores for the six emotional categories: anger, fear, happiness, neutrality, sadness, and surprise. This structure enables efficient access to each user's personality and emotion data.

For example, in the ENFJ folder, CSV files for all CMACD users with the ENFJ personality type are stored. The "posts" column records the content of each user's posts, with 50 rows representing 50 Weibo posts per user. Other columns provide multi-label emotion annotations, with each post assigned corresponding emotion scores.

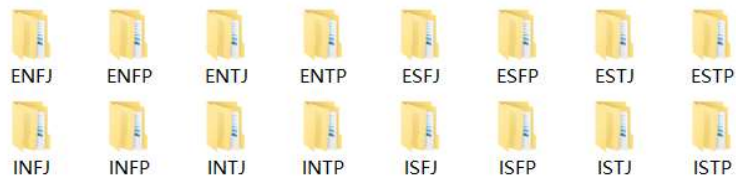

Fig. 12.Storage Structure of the CMACD Dataset.

Fig. 13 Annotated segments with emotional labels and intensity scores. Fig. 13 shows an example of a user's post with annotated emotional segments and their corresponding energy-level scores. This Fig. 13 illustrates a sample from the CMACD dataset, where each user's post is accompanied by emotional segments, each with a quantifiable intensity score. Each post is followed by six emotional categories: anger, fear, happiness, neutrality, sadness, and surprise, with each emotion assigned a measurable intensity score. These annotations provide a detailed insight into the emotional content of each post, reflecting the depth of emotional expression within the user's content.

## Technical Validation

To validate the usability and effectiveness of the CMACD dataset, we conducted personality classification and emotion classification tests using various classic algorithms and deep learning models. For the sake of experimental reproducibility, the algorithms or models employed standard or basic parameter settings without any special optimization. Additionally, we performed Pearson correlation analysis using the vector values corresponding to the four dimensions of the 16 personality labels to examine the interrelationships between the four aspects (axes) of the MBTI personality types, thus assessing the rationality of the dataset distribution. Furthermore, based on the annotated emotion intensity scores in the dataset, we performed Pearson correlation analysis to explore the relationships between the six emotions, observing whether the correlations between the emotions align with human cognitive understanding of emotional relationships.

**1. Validation of MBTI Personality Types**

The validation of MBTI personality types encompasses five key components: data preparation, selected algorithms and models, experimental protocols, experimental results, and testing the correlations across the four MBTI axes.

**Data Preparation.** To objectively validate the dataset, the following steps were undertaken:

**(1) Removal of Personality Keywords:** To prevent the interference of MBTI personality keywords during training, we excluded these terms from the dataset, including all uppercase and lowercase variations. The 16 personality type keywords are as follows: ['INFJ', 'ENTP', 'INTP', 'INTJ', 'ENTJ', 'ENFJ', 'INFP', 'ENFP', 'ISFP', 'ISTP', 'ISFJ', 'ISTJ', 'ESTP', 'ESFP', 'ESTJ', 'ESFJ'].

**(2) Sentence Integrity:** Unlike traditional machine semantic understanding, this project involves evaluating human language behavior that reflects individual personality traits. Therefore, we maintained the integrity of each sentence used for training. Stopwords were not utilized in the data processing.

**(3) Segmentation Tools:** In this experiment, we employed the Chinese word segmentation tool "jieba" to process Chinese words and phrases into logical segments that align with the structure of the Chinese language. Unlike English, where word segmentation is typically achieved by spaces between words, Chinese segmentation is more complex. For example, "apple" in Chinese is "苹果" (píng guǒ), and it is more intuitive to segment it as "苹果" rather than breaking it into the characters "苹" and "果." A sample of segmentation using "jieba" is shown in Fig. 14.

原文： 种植车厘子的方法：1、处理植株在种植车厘子前，要先将果树根部的烂根、枯根以及病害根剪掉，再向其喷洒除菌药剂进行消毒杀菌，然后将车厘子果树，按照每隔40厘米的间距，栽到土壤中，最后为植株每3天浇一次水。

分词结果： 种植 车厘子 的 方法 ： 1 、 处理 植株 在 种植 车厘子 前 ， 要 先 将 果树 根部 的 烂根 、 枯根 以及 病害 根 剪掉 ， 再 向 其 喷洒 除菌 药剂 进行 消毒 杀菌 ， 然后 将 车 厘子 果树 ， 按照 每隔 40 厘米 的 间距 ， 栽种 到 土壤 中 ， 最后 为 植株 每 3 天 浇 一次 水 。

Fig. 14 Examples of using jieba word segmentation.

**(4) Generation of word embeddings or use of Tfidf to create the TFIDF matrix.**
Term Frequency-Inverse Document Frequency (TF-IDF) is a classical and widely used method in natural language processing (NLP). The fundamental concept is as follows: if a word appears frequently in a document but infrequently across other documents, then the word is considered highly important for that specific document. TFIDF transforms textual corpora into feature vectors represented by words. This process involves vectorizing the database posts into a matrix of token counts for the model. In this experiment, we set the parameters `max_df=0.97` (removing words that appear in more than 97% of the samples) and `min_df=0.03` (removing words that appear in less than 3% of the samples). The maximum number of features was set to 'max_features=5000`.

**Selected Algorithm Models**
To validate the usability and adaptability of the dataset, this project employs a range of algorithms based on the dataset. The following models are chosen for the classification and prediction of 16 personality types: Logistic Regression (Logit), AdaBoost, Gaussian Naive Bayes (GNB), Random Forest, Decision Tree, Support Vector Machine (SVM), K-Nearest Neighbors (KNN), eXtreme Gradient Boosting (XGBoost), Naive Bayes, Multilayer Perceptron (MLP), Recurrent Neural Network (RNN), and Bidirectional Encoder

Representations from Transformers (BERT[42]).

**Experimental Design**. Based on the aforementioned models, the experimental design considers the different input methods for MBTI display, large models like BERT, and other models. The specific approach is as follows:

(1) In accordance with the MBTI classification rules, the experiment transforms the 16-class classification problem into four binary classification tasks, mapped as follows: {'I': 0, 'E': 1, 'N': 0, 'S': 1, 'F': 0, 'T': 1, 'J': 0, 'P': 1}.

(2) The BERT model utilizes word embeddings as input, while other models or methods use the TF-IDF matrix as input.

**Experimental Results.** Based on the algorithms, models, and experimental design outlined above, the personality type dataset is divided into training and test sets in a 7:3 ratio for training and validation. The final experimental results (Accuracy) are shown in Table 3.

| Models | I/E | N/S | T/F | J/P |
| --- | --- | --- | --- | --- |
| Logit | 0.6286 | 0.7054 | 0.7100 | 0.6124 |
| AdaBoost | 0.6188 | 0.7241 | 0.7183 | 0,5701 |
| GNB | 0.5465 | 0.5909 | 0.5919 | 0.5374 |
| RandomForest | 0.6253 | 0.7251 | 0.7196 | 0.5799 |
| DecisionTree | 0.5454 | 0.6145 | 0.6116 | 0.5242 |
| SVM | 0.6254 | 0.7054 | 0.7110 | 0.5940 |
| KNN | 0.62.90 | 0.7257 | 0.7206 | 0.5871 |
| XGBoost | 0.6344 | 0.7142 | 0.6902 | 0.5971 |
| Naive Bayes | 0.5726 | 0.5940 | 0.6030 | 0.5499 |
| Bagging | 0.6244 | 0.7041 | 0.7074 | 0.5962 |
| MLP | 0.6191 | 0.7241 | 0.7195 | 0.5745 |
| RNN | 0.6290 | 0.7257 | 0.7206 | 0.5871 |
| BERT | **0.7674** | **0.7851** | **0.7963** | **0.7388** |

**Table 3. Experimental results (Accuracy) of the 16 personality dataset across multiple classical algorithms and models.**

The results indicate that classical algorithms, neural networks, and large language models exhibit varying classification capabilities on our dataset, thereby validating the effectiveness and adaptability of the dataset for personality classification. Notably, BERT demonstrated exceptional classification performance, highlighting the unique advantages of large language models. The results of the MBTI text classification model based on BERT can serve as a benchmark for the classification performance of this dataset.

**MBTI Axis Correlation Test**. According to MBTI theory, the 16 personality labels are composed of four different letters, each representing one axis of the four-dimensional framework. Each of these axes has two polarities, which can be represented in binary form (for example, the I/E axis, where I is 1 and E is 0). Consequently, each label can be expressed as a 4-dimensional vector based on the polarities of the axes. For instance, the personality type "INFJ" can be represented as the vector [1,1,0,1]. Using the vector values corresponding to the four axes of the MBTI labels in the dataset, we calculated the Pearson correlation coefficient to analyze the relationships between the four axes. The resulting correlation heatmap is shown in Fig. 15.

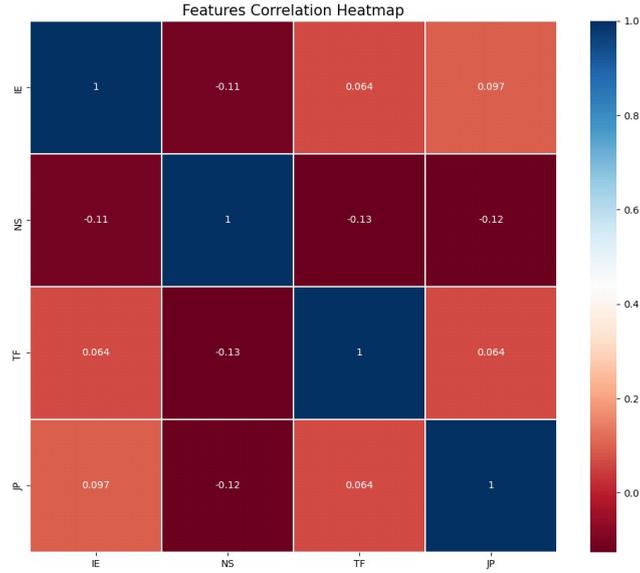

Fig. 15.MBTI four-axis correlation heat map. The absolute value of the Pearson correlation coefficient indicates the strength of the correlation. The closer the correlation coefficient is to 1 or -1, the stronger the correlation; conversely, the closer it is to 0, the weaker the correlation. As shown in Fig. 15, the correlations between the four axes are generally weak, indicating that the personality traits on each axis are relatively independent in our dataset. This suggests that the distribution of the four axes in our dataset is reasonable.

---

**Validation of Six Emotion Labels**

To validate the quality of the six emotion annotations in the CMACD dataset, we employed three methods to assess usability, accuracy, and the reasonableness of the overall label distribution:

**(1) Multi-label Classification Experiment to Assess Dataset Usability**

We trained and evaluated our emotion dataset using Support Vector Machine (SVM), FastText, TextCNN, TextRNN, and BERT models. To validate the usability and effectiveness of the dataset, the experiment was conducted using the standard or baseline parameter settings for each algorithm or model, without any special optimization. Two evaluation metrics were used:

$E_1\_Acc$: Accuracy of the label with the highest energy score.

$E_x\_Acc$: Accuracy of labels with non-maximum energy scores.

The training and evaluation results are shown in Table 4.

| Models | $E_1\_Acc$ | $E_x\_Acc$ |
|---|---|---|
| SVM | 0.6577 | 0.6213 |
| FastText | 0.7245 | 0.6857 |
| TextCNN | 0.7336 | 0.7089 |
| TextRNN | 0.6989 | 0.6711 |
| BERT | 0.9284 | 0,8865 |

**Table 4.Multi-label Emotion Classification Evaluation Results.**

The experimental results above indicate that, although different algorithms and models exhibit varying classification performance based on the CMACD dataset, the samples in the dataset possess the necessary features for classification, thus validating the dataset's usability. Among the results, the BERT model achieved the best performance, and its results can serve as the benchmark for emotion classification on this dataset.

**(2) Manual Spot Check to Assess the Accuracy of Emotion Annotations**

To comprehensively evaluate the quality of the AI-based machine learning automatic annotations in the emotion dataset, a traditional manual spot check was conducted to assess the quality of non-human-supervised annotations.

Manual Annotation Rules: Voting Method. A team of three annotators (each annotating posts with a single label) determined the final label based on a voting principle.

Method: A random sample of 1,000 posts was selected from the dataset, and all labels previously assigned by the annotation tool were removed. These 1,000 samples were then manually annotated with single labels.

Result Comparison: The manually annotated 1,000 samples (single-label) were compared with the original multi-label results assigned by the annotation tool, which included energy scores. The hit rate for the top energy label (the label with the highest emotion energy score in a sentence) was 83.1%, while the hit rate for the top two energy labels (the highest and the second-highest emotion energy labels in a sentence) was 92.3%. This result validates that our emotion dataset has a high level of accuracy, and it also demonstrates that the full-label method in EQN is beneficial for capturing micro-emotions.

**(3) Evaluation of the Reasonableness of the Overall Distribution of Six Emotion Labels**

To observe the correlations between the six annotated emotions, we used the Pearson correlation coefficient to calculate and analyze the relationships between the emotions based on the quantified emotion data. This approach allows us to assess whether the overall distribution of the labels is reasonable and provides an indication of the overall annotation accuracy within the dataset. The resulting correlation heatmap is shown in Fig. 16.

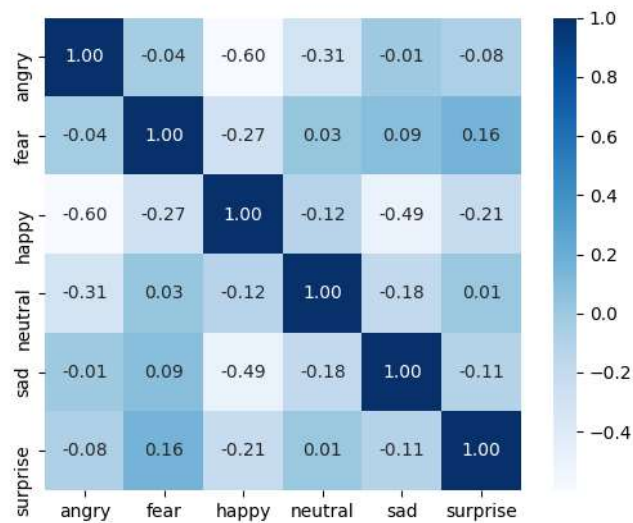

Fig. 16.Correlation Heatmap of Six Emotions and Quantitative Annotations. As shown in the Fig. 16, emotions such as "angry" and "happy," as well as "happy" and "sad," exhibit a strong negative correlation, while "fear" and "surprise" show a moderate positive correlation. These correlations between emotions are largely consistent with the expression of emotions in real-life scenarios. This also indicates that the distribution of emotion samples in our dataset, along with the overall accuracy of the annotations, is relatively high.

Emotion and personality are two closely intertwined concepts of human psychological states. Emotion represents an individual's subjective experience, while personality refers to relatively stable patterns of behavior and cognition. This paper, based on a large Chinese microblogging platform, strictly selected 11,338 valid users from over 50,000 publicly labeled users with MBTI personality traits. It introduces the CMACD dataset, the first dataset that integrates personality and emotion with intensity labels for affective computing. Through the use of feature engineering, neural networks, RNN, BERT, and other algorithms and models, the usability and effectiveness of the CMACD dataset were validated. This dataset provides robust data support for further research on the relationship between personality and emotion, or for quantifying human behavior from a psychological perspective, as well as applications in AI-driven behavioral prediction, fake information detection, human-machine alignment, humanoid robots, and more.

## Usage Notes

Although this dataset has undergone privacy protection measures, it involves research on human personality and emotions. To ensure user safety, CMACD is made available free of charge only to researchers with a legitimate need. Researchers wishing to use CMACD can apply by emailing: annezjy94@163.com.

To demonstrate the features and application value of this dataset, and to facilitate basic testing and feedback, we have made a small sample of CMACD publicly available. The small sample dataset is now accessible at: https://github.com/yeaso/Chinese-Affective-Computing-Dataset

## Code Availability

To demonstrate the features and application value of this dataset, and to facilitate basic testing and feedback, we have made a small sample of CMACD publicly available. The small sample dataset is now accessible at: https://github.com/yeaso/Chinese-Affective-Computing-Dataset

## Acknowledgements

This research is partially supported by the 242 National InformationSecurity Projects, PR China under Grant 2020A065.

## Author contributions

Jingyi Zhou: Conceptualization, Codes, Methodology, Writing original draft.
Senlin Luo: Project administration, Supervision, Writing – review & editing.
Haofan Chen: Resources, Testing, Codes – review & editing.

## Competing interests

The authors declare that they have no known competing financial
interests or personal relationships that could have appeared to influence
the work reported in this paper.

## References

1.Malhotra, A. & Jindal, R. XAI Transformer based Approach for Interpreting Depressed and Suicidal User Behavior on Online Social Networks. *Cognitive Systems Research*.**84**, 101186 (2024).

2.Munoz, S. & Iglesias, C. a. Detection of the Severity Level of Depression Signs in Text Combining a Feature-Based Framework with Distributional Representations.*APPLIED SCIENCES-BASEL*.**13**, (2023).

3.Priya, P., Firdaus, M. &Ekbal, A. A multi-task learning framework for politeness and emotion detection in dialogues for mental health counselling and legal aid.*EXPERT SYSTEMS WITH APPLICATIONS*.**224**, (2023).

4.Bussoletti, M., Castro, D., Zebdi, R. &MatarTouma, V. Prevalence of depression and protective factors in a population of children aged 8 to 10 years, suffering from specific learning disorders, in a special education and home care service (SESSAD). *L'Encephale*.**50**, 400–405 (2024).

5.Santhosh, J., Pai, A. &Ishimaru, S. Toward an Interactive Reading Experience: Deep Learning Insights and Visual Narratives of Engagement and Emotion. *IEEE ACCESS*.**12**, 6001–6016 (2024).

6.Martínez-España, R. et al. Methodology for Measuring Individual Affective Polarization Using Sentiment Analysis in Social Networks. *IEEE ACCESS*.**12**, 102035–102049 (2024).

7.Zhao, Y., Ma, J. & Chow, T. Extractive Negative Opinion Summarization of Consumer Electronics Reviews.*IEEE TRANSACTIONS ON CONSUMER ELECTRONICS*.**70**, 3521–3528 (2024).

8.Lin, W. & Liao, L. Lexicon-based prompt for financial dimensional sentiment analysis. *EXPERT SYSTEMS WITH APPLICATIONS*.**244**, (2024).

9Kaplan, H., Weichselbraun, A. &Brasoveanu, A. Integrating Economic Theory, Domain Knowledge, and Social Knowledge into Hybrid Sentiment Models for Predicting Crude Oil Markets.*COGNITIVE COMPUTATION*.**15**, 1355–1371 (2023).

10.Holliday, D., Iyengar, S., Lelkes, Y. & Westwood, S. Uncommon and nonpartisan: Antidemocratic attitudes in the American public. *PROCEEDINGS OF THE NATIONAL ACADEMY OF SCIENCES OF THE UNITED STATES OF AMERICA*.**121**, (2024).

14.Zhou, L., Gao, J., Li, D. & Shum, H.-Y. The Design and Implementation of XiaoIce, an Empathetic Social Chatbot.*Computational Linguistics*.**46**, 53–93 (2020).

15.Li, Y. et al. Towards an Online Empathetic Chatbot with Emotion Causes. Preprint at https://doi.org/10.1145/3404835.3463042 (2021).

11.Luna-Jiménez, C., Gil-Martín, M., D'Haro, L., Fernández-Martínez, F. & San-Segundo, R. Evaluating emotional and subjective responses in synthetic art-related dialogues: A multi-stage framework with large language models. *EXPERT SYSTEMS WITH APPLICATIONS*.**255**, (2024).

12.Tran, Q. et al. Robustness Analysis uncovers Language Proficiency Bias in Emotion Recognition Systems.*2023 11TH INTERNATIONAL CONFERENCE ON AFFECTIVE COMPUTING AND INTELLIGENT INTERACTION(ACII)*.doi:10.1109/ACIIW59127.2023.10388123,(2023).

13.Mao, R., Liu, Q., He, K., Li, W. & Cambria, E. The Biases of Pre-Trained Language Models: An Empirical Study on Prompt-Based Sentiment Analysis and Emotion Detection. *IEEE TRANSACTIONS ON AFFECTIVE COMPUTING*.**14**, 1743–1753 (2023).

16.Nagamanjula, R., &Pethalakshmi, A. A novel framework based on bi-objective optimization and LAN2FIS for Twitter sentiment analysis. *Social Network Analysis and Mining*.**10**, (2020).

17.Lukas Christ, ShahinAmiriparian, Alice Baird, Panagiotis Tzirakis, Alexander Kathan, Niklas Müller, Lukas Stappen, Eva-Maria Meßner, Andreas König, Alan Cowen, Erik Cambria, and Björn W. Schuller.   The MuSe 2022 Multimodal Sentiment Analysis Challenge: Humor, Emotional Reactions, and Stress. *In Proceedings of the 3rd International on Multimodal Sentiment Analysis Workshop and Challenge (MuSe' 22)*. Association for Computing Machinery, New York, NY, USA, 5–14,https://doi.org/10.1145/3551876.3554817,(2022).

18.Lung-Hao Lee, Jian-Hong Li and Liang-Chih Yu, "Chinese EmoBank: Building Valence-Arousal Resources for Dimensional Sentiment Analysis," ACM Trans. Asian and Low-Resource Language Information Processing, vol. 21, no. 4, article 65, (2022).

19.Demszky, D. et al. GoEmotions: A Dataset of Fine-Grained Emotions. *InProceedingsofthe 58th Annual Meeting of the Association for Computational Linguistics*.**372**,4040–4054.doi:10.18653/v1/2020.acl-main,(2020).

20. Jingyi Zhou, Senlin Luo, Haofan Chen. Expansion Quantization Network: An Efficient Micro-emotion Annotation and Detection Framework. Preprint at https://doi.org/10.48550/arXiv.2411.06160 (2024).

21.PENGDanling. General Psychology (Revised Edition).*Beijing Normal University Press*, (2001).

22.Kashani, A., Pfau, J., El-Nasr, M., & IEEE. Assessing the Impact of Personality on Affective States from Video Game Communication.*2023 11TH INTERNATIONAL CONFERENCE ON AFFECTIVE COMPUTING AND INTELLIGENT INTERACTION WORKSHOPS AND DEMOS,*doi:10.1109/ACIIW59127.2023.10388185, (2023).

23.Kamran, S. et al. EmoDNN: understanding emotions from short texts through a deep neural network ensemble.*NEURAL COMPUTING & APPLICATIONS*.**35**, 13565–13582 (2023).

24.Kitkowska, A., Shulman, Y., Martucci, L. &Wätlund, E. Designing for privacy: Exploring the influence of affect and individual characteristics on users' interactions with privacy policies. *COMPUTERS & SECURITY*.**134**, (2023).

25.Amin, M., Cambria, E. & Schuller, B. Can ChatGPT's Responses Boost Traditional Natural Language Processing? *IEEE INTELLIGENT SYSTEMS*.**38**, 5–11 (2023).

26."International Mother Language Day": Exploring the challenges and new approaches to Chinese teaching and communication,*United Nations News,* https://news.un.org/zh/audio/2017/02/308982, (2017).

27.The number of Internet users in China has reached 1.092 billion,*Chinese government website*,https://www.gov.cn/yaowen/liebiao/202403/content_6940952.htm, (2024).

28.2023 Weibo Young User Development Report,*Weibo Data Center,*https://data.weibo.com/report/reportDetail?id=471, (2024).

29.WangHao, Gong Lijuan, Zhou Zeyu, Fan Tao, & Wang Yongsheng. Detecting Mis/Dis-information from Social Media with Semantic Enhancement.*Data Analysis and Knowledge Discovery*.**7**, 48–60 (2023).

30.WangYouwei, Feng Lizhou, Wang Weiqi, &HouYudong. Weibo Rumor Detection Based on Heterogeneous Graph of Event-Word-Feature.*Journal of Chinese Information Processing*.**37**, 161–174 (2023).

31.Kong, D. et al. Public Discourse and Sentiment Toward Dementia on Chinese Social Media: Machine Learning Analysis of Weibo Posts. *JOURNAL OF MEDICAL INTERNET RESEARCH*.**24**, (2022).

32.Tan, M., Wu, Z., Li, J., Liang, Y. &Lv, W. Analyzing the impact of unemployment on mental health among Chinese university graduates: a study of emotional and linguistic patterns on Weibo.*FRONTIERS IN PUBLIC HEALTH*.**12**, (2024).

33.LiTiejun, Yan Duanwu, & Yang Xiongfei. Recommending Microblogs Based on Emotion-Weighted Association Rules.*Data Analysis and Knowledge Discovery*.**4**, 27–33 (2020).

34.Xiong, W. & Zhang, Y. An intelligent film recommender system based on emotional analysis.*PEERJ COMPUTER SCIENCE*.**9**, (2023).

35Liu, J., Liu, W., Yan, C. & Liu, X. Study on the Temporal and Spatial Evolution Characteristics of Chinese Public's Cognition and Attitude to 'Double Reduction' Policy Based on Big Data. *BIG DATA RESEARCH*.**34**, (2023).

36. YangYunjia & Shi Xingmin. Social Sensing and Spatio-temporal Pattern of 'Elephants Wandering North' Event Based on Micro-blog Big Data. *Geography and Geo-information Science*. **39**, 10–16 (2023).

37. Ma, M. et al. Analysis of public emotion on flood disasters in southern China in 2020 based on social media data. *NATURAL HAZARDS*. **118**, 1013–1033 (2023).

38. Gao, H., Guo, D., Wu, J. & Li, L. Weibo Users' Emotion and Sentiment Orientation in Traditional Chinese Medicine (TCM) During the COVID-19 Pandemic. DISASTER MEDICINE AND PUBLIC HEALTH PREPAREDNESS 16, 1835–1838 (2022).

39. Liu, Y., Liu, S., Ye, D., Tang, H. & Wang, F. Dynamic impact of negative public sentiment on agricultural product prices during COVID-19. *JOURNAL OF RETAILING AND CONSUMER SERVICES*. **64**, (2022).

40. Yu, S., Eisenman, D. & Han, Z. Temporal Dynamics of Public Emotions During the COVID-19 Pandemic at the Epicenter of the Outbreak: Sentiment Analysis of Weibo Posts From Wuhan. *JOURNAL OF MEDICAL INTERNET RESEARCH*. **23**, (2021).

41. The Evaluation of Weibo Emotion Classification Technology, *SMP2020-EWECT*. https://smp2020ewect.github.io/, (2020).

42. V. S. Kodiyala and R. E. Mercer, "Emotion Recognition and Sentiment Classification using BERT with Data Augmentation and Emotion Lexicon Enrichment," 2021 20th IEEE International Conference on Machine Learning and Applications (ICMLA), Pasadena, CA, USA, 2021, pp. 191-198, doi: 10.1109/ICMLA52953.2021.00037.